\pdfoutput=1

\documentclass[11pt]{article}

\usepackage[table,xcdraw]{xcolor}
\usepackage{acl}
\usepackage{float}
\usepackage{hyperref}
\usepackage{amsmath} 
\usepackage{amssymb} 
\usepackage{cleveref}
\usepackage{graphicx} 
\usepackage{natbib} 
\usepackage{booktabs} 
\usepackage{array}    %

\usepackage{times}
\usepackage{latexsym}
\usepackage{booktabs}
 \usepackage{amsmath} 
\usepackage[skins]{tcolorbox} 
\usepackage{enumitem}    
\usepackage{geometry}  

\usepackage{multirow}
\usepackage{amssymb}
\usepackage{pifont}
\usepackage{xspace}
\newcommand*{\yoruba}{Yor\`ub\'a\xspace}

\definecolor{bblue}{HTML}{4F81BD}
\definecolor{rred}{HTML}{C0504D}
\definecolor{ggreen}{HTML}{9BBB59}
\usepackage{times}
\usepackage{latexsym}

\usepackage[T1]{fontenc}

\usepackage[utf8]{inputenc}

\usepackage{microtype}

\usepackage{inconsolata}

\usepackage{graphicx}
\usepackage{cleveref}
\usepackage[utf8]{inputenc}
\usepackage{csquotes}

\usepackage{times}
\usepackage{latexsym}
\usepackage{booktabs}
\usepackage{microtype}
\usepackage{amssymb}
\usepackage{pifont}
\usepackage{graphicx}
\usepackage{placeins}
\usepackage{natbib}
\usepackage{tablefootnote}
\usepackage{longtable}
\usepackage{cleveref}
\usepackage{enumitem}
\usepackage{booktabs}
\usepackage{pdfpages}

\usepackage{tabularx}
\usepackage{multirow}
\usepackage{xspace}

\usepackage{linguex}
\alignSubExtrue

\newcommand\datasetname{\textcolor{black}{\textsc{Brighter}}}

\usepackage[verbose]{newunicodechar}

\usepackage{times}
\usepackage{latexsym}
\usepackage{microtype}
\usepackage{amssymb}
\usepackage{pifont}
\usepackage{graphicx}
\usepackage{placeins}
\usepackage{natbib}
\usepackage{tablefootnote}
\usepackage{longtable}
\usepackage{cleveref}
\usepackage{enumitem}
\usepackage{booktabs}
\usepackage{pdfpages}
\usepackage[T1]{fontenc}

\usepackage{tabularx}
\usepackage{multirow}
\usepackage{xspace}

\usepackage[textsize=footnotesize]{todonotes}
\usepackage{linguex}
\usepackage{subcaption}
\alignSubExtrue
\usepackage[verbose]{newunicodechar}

\usepackage{inconsolata}

\usepackage{makecell}
\global

\tcbset{
  aibox/.style={
    width=474.18663pt,
    top=10pt,
    colback=white,
    colframe=black,
    colbacktitle=black,
    enhanced,
    center,
    attach boxed title to top left={yshift=-0.1in,xshift=0.15in},
    boxed title style={boxrule=0pt,colframe=white,},
  }
}
\newtcolorbox{AIbox}[2][]{aibox,title=#2,#1}
\newcommand{\bestmono}{\cellcolor{blue!30}}
\newcommand{\bestcross}{\cellcolor{orange!20}}
\newcommand{\bestmlm}{\cellcolor{blue!30}}
\newcommand{\bestllm}{\cellcolor{orange!30}}

\author{
Shamsuddeen Hassan Muhammad$^{1,2}$\thanks{Equal contribution}, Nedjma Ousidhoum$^{3*}$,
Idris Abdulmumin$^4$, \\
\bf Jan Philip Wahle$^5$, Terry Ruas$^5$, Meriem Beloucif$^6$, Christine de Kock$^7$, \\
\bf Nirmal Surange$^8$, Daniela Teodorescu$^9$, Ibrahim Said Ahmad$^{10}$, \\
\bf David Ifeoluwa Adelani$^{11,12,13}$, Alham Fikri Aji$^{14}$, Felermino D. M. A. Ali$^{15}$, \\
\bf Ilseyar Alimova$^{31}$, Vladimir Araujo$^{16}$, Nikolay Babakov$^{17}$, Naomi Baes$^{7}$,\\
\bf Ana-Maria Bucur$^{18,19}$, Andiswa Bukula$^{20}$, Guanqun Cao$^{21}$, Rodrigo Tufi\~no$^{22}$, \\
\bf Rendi Chevi$^{14}$, Chiamaka Ijeoma Chukwuneke$^{23}$, Alexandra Ciobotaru$^{18}$, \\
\bf Daryna Dementieva$^{24}$, Murja Sani Gadanya$^{2}$, Robert Geislinger$^{25}$, Bela Gipp$^5$, \\
\bf Oumaima Hourrane$^{26}$, Oana Ignat$^{27}$, Falalu Ibrahim Lawan$^{28}$, Rooweither Mabuya$^{20}$, \\
\bf Rahmad Mahendra$^{29}$, Vukosi Marivate$^{4,30}$, Alexander Panchenko$^{31,32}$, Andrew Piper$^{12}$, \\
\bf Charles Henrique Porto Ferreira$^{33}$, Vitaly Protasov$^{32}$, Samuel Rutunda$^{34}$, \\
\bf Manish Shrivastava$^{8}$, Aura Cristina Udrea$^{35}$, Lilian Diana Awuor Wanzare$^{36}$, Sophie Wu$^{12}$, \\
\bf Florian Valentin Wunderlich$^5$, \bf Hanif Muhammad Zhafran$^{37}$, Tianhui Zhang$^{38}$, Yi Zhou$^3$, \\
\bf Saif M. Mohammad$^{39}$ \\
\footnotesize $^{1}$Imperial College London, $^{2}$Bayero University Kano, $^3$Cardiff University,\\
\footnotesize $^{4}$Data Science for Social Impact, University of Pretoria, $^{5}$University of Göttingen,
$^{6}$Uppsala University,\\
\footnotesize  $^{7}$University of Melbourne, $^{8}$IIIT Hyderabad, $^{9}$University of Alberta, $^{10}$Northeastern University, $^{11}$MILA, $^{12}$McGill University,\\
\footnotesize $^{13}$Canada CIFAR AI Chair, $^{14}$MBZUAI, $^{15}$LIACC, FEUP, University of Porto, $^{16}$Sailplane AI,\\
\footnotesize $^{17}$University of Santiago de Compostela, $^{18}$University of Bucharest, $^{19}$Universitat Politècnica de València, $^{20}$SADiLaR,\\
\footnotesize $^{21}$University of York, $^{22}$Universidad Politécnica Salesiana, $^{23}$Lancaster University, $^{24}$Technical University of Munich,\\
\footnotesize $^{25}$Hamburg University, $^{26}$Al Akhawayn University, $^{27}$Santa Clara University, $^{28}$Kaduna State University, $^{29}$Universitas Indonesia,\\
\footnotesize $^{30}$Lelapa AI, $^{31}$Skoltech, $^{32}$AIRI, $^{33}$Centro Universitário FEI, $^{34}$Digital Umuganda,\\
\footnotesize $^{35}$National University of Science and Technology Politehnica Bucharest, $^{36}$Maseno University, $^{37}$Institut Teknologi Bandung,\\
\footnotesize $^{38}$University of Liverpool, $^{39}$National Research Council Canada\\
\footnotesize{\texttt{Contact: s.muhammad@imperial.ac.uk, OusidhoumN@cardiff.ac.uk}
}
}

\title{\datasetname: BRIdging the Gap in Human-Annotated Textual Emotion Recognition Datasets for 28 Languages}

\begin{document}

\maketitle
\begin{abstract}
People worldwide use language in subtle and complex ways to express emotions. Although emotion recognition--an umbrella term for several NLP tasks--impacts various applications within NLP and beyond, most work in this area has focused on high-resource languages. This has led to significant disparities in research efforts and proposed solutions, particularly for under-resourced languages, which often lack high-quality annotated datasets.
In this paper, we present \datasetname--a collection of multi-labeled, emotion-annotated datasets in 28 different languages and across several domains. \datasetname~primarily covers low-resource languages from Africa, Asia, Eastern Europe, and Latin America, with instances labeled by fluent speakers. We highlight the challenges related to the data collection and annotation processes, and then report experimental results for monolingual and crosslingual multi-label emotion identification, as well as emotion intensity recognition. We analyse the variability in performance across languages and text domains, both with and without the use of LLMs, and show that the \datasetname~ datasets represent a meaningful step towards addressing the gap in text-based emotion recognition.
\end{abstract}

\begin{figure}[h]
    \centering
     \scalebox{0.8}{
        \includegraphics[trim=0 0 0 40, clip, width=1.25\linewidth]{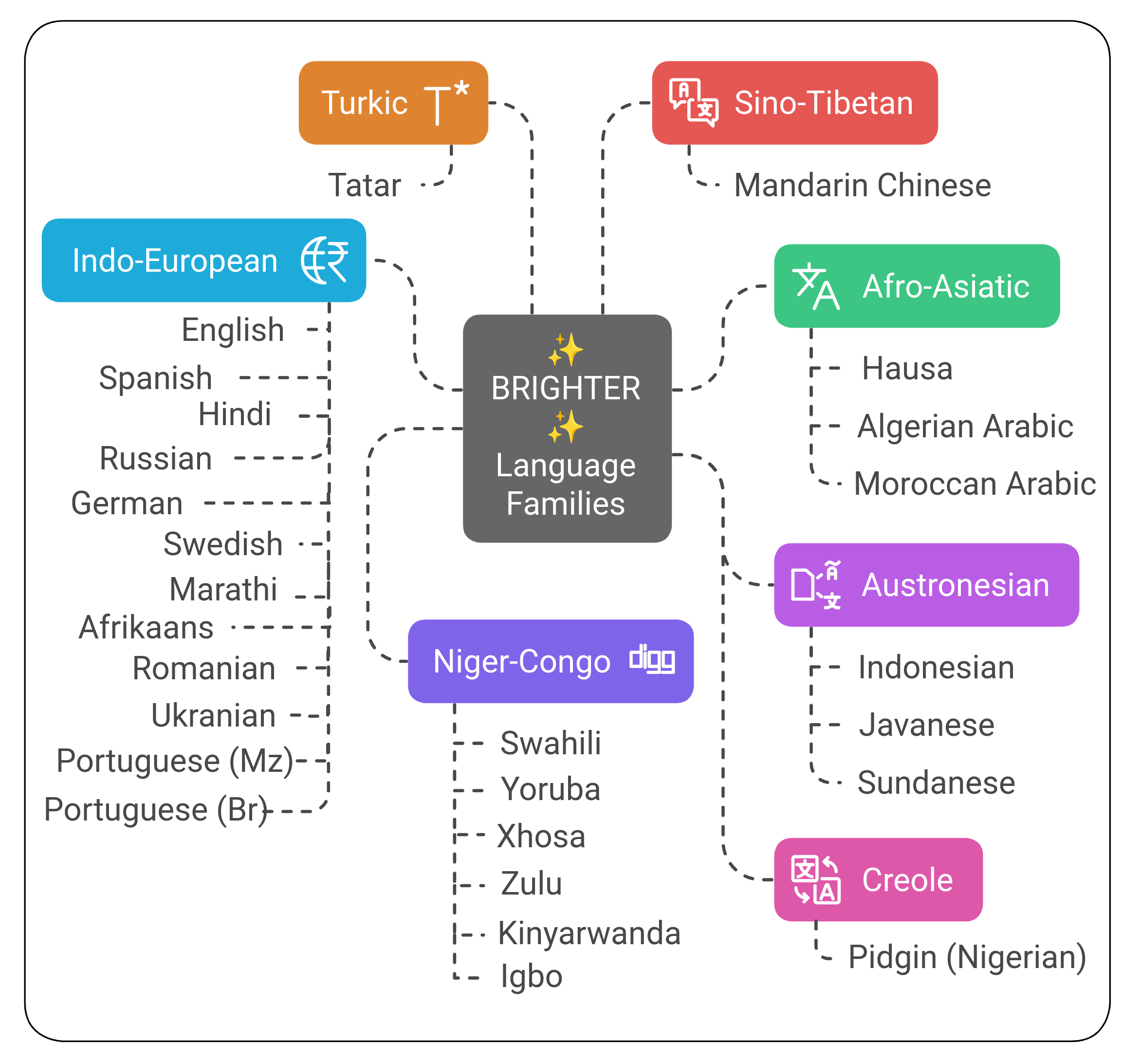}
    }

    \caption{\textbf{Languages included in  \datasetname} and their language families.}

    \label{fig:lgge_fams}
\end{figure}

\begin{figure*}[h]
    \centering
    \scalebox{0.8}{
    \includegraphics[width=\linewidth]
    {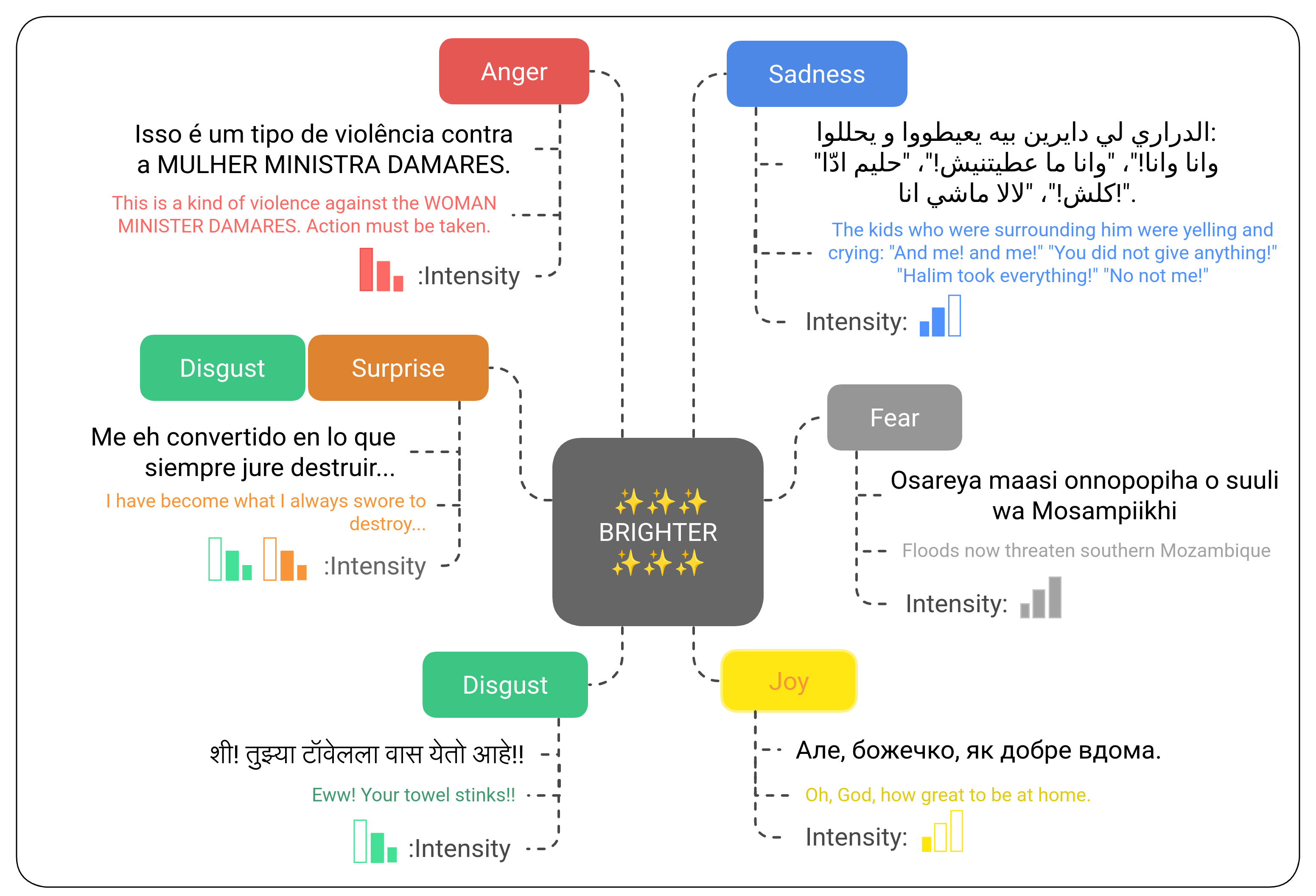}
    }
    \caption{Examples from the \datasetname~dataset collection in 6 different languages with their translations and intensity levels. Note that the instances can have one or more labels (e.g., disgust and surprise as shown in the figure).} 
    \label{fig:brighter_examples}
\end{figure*}

\section{Introduction}

While emotions are expressed and managed daily, they are complex, nuanced, and sometimes hard to articulate and interpret. That is, people use language in subtle and complex ways to express emotions across languages and cultures \cite{wiebe2005annotating,mohammad-kiritchenko-2018-understanding,mohammad2018semeval} and perceive them subjectively, even within the same culture or social group. 
Emotion recognition is at the core of several NLP applications in healthcare, dialogue systems, computational social science, digital humanities, narrative analysis, and many others \cite{mohammad-etal-2018-semeval,saffar2023textual}. It is an umbrella term for multiple NLP tasks, such as detecting the possible emotions of the speaker, identifying what emotion a piece of text is conveying, and detecting the emotions evoked in a reader \cite{mohammad2022ethics}. In this paper, we use \textit{emotion recognition} to refer to \textit{perceived} emotions, i.e., what emotion most people think the speaker might have felt given a sentence or a short text snippet uttered by them.

Most work on emotion recognition has focused on high-resource languages such as English, Spanish, German, and Arabic \cite{strapparava-mihalcea-2007-semeval,seyeditabari2018emotion,chatterjee-etal-2019-semeval,kumar2022discovering}. This is partly due to the unavailability of datasets in under-served languages, which has led to a major research gap in the area, which is particularly noticeable in low-resource languages. That is, despite the linguistic diversity present in different parts of the world, such as Africa and Asia, which are home to more than 4,000 languages\footnote{\url{https://www.ethnologue.com/insights/how-many-languages}}, few emotion recognition resources are available in these languages. To bridge this gap, we introduce \datasetname--a collection of manually annotated emotion datasets for 28 languages containing nearly 100,000 instances from diverse data sources: speeches, social media, news, literature, and reviews. The languages belong to 7 language families (see Figure \ref{fig:lgge_fams}) and are predominantly low-resource, mainly spoken in \textbf{Africa}, \textbf{Asia}, \textbf{Eastern Europe}, \textbf{Latin America}, along with mid- to high-resource languages such as English. Each instance in \datasetname~is curated and annotated by fluent speakers based on six emotion classes: \textit{joy, sadness, anger, fear, surprise, disgust,} and none for neutral. The instances are multi-labeled and include 4 levels of intensity that vary from 0 to 3 (examples in Figure \ref{fig:brighter_examples}).
We describe the collection, annotation, and quality control steps used to construct \datasetname. We then test various baseline experiments and observe that LLMs still struggle with recognising perceived emotions in text. We further report on the observed discrepancies across languages such as the fact that, for low-resource languages, LLMs perform significantly better when prompted in English.
We make our datasets public\footnote{The datasets are available at \url{https://brighter-dataset.github.io}. Note that they were used in SemEval-2025 Task 11, which attracted over 700 participants~\cite{muhammad-etal-2025-semeval}.}, which presents an important step towards work on emotion recognition and related tasks as we involve local communities in the collection and annotation. Our insights into language-specific characteristics of emotions in text, nuances, and challenges may enable the creation of more inclusive digital tools.

\section{The \datasetname ~Dataset Collection}
\begin{table*}[ht]
\small
\centering
    \resizebox{\linewidth}{!}{
    \begin{tabular}{p{4.1cm}p{2.4cm}p{1.5cm}p{1cm}p{0.8cm}p{0.8cm}p{0.8cm}p{0.8cm}}
    \toprule
    \textbf{Language} & \textbf{Data source(s)} & \textbf{\#Annotators (total)} & \textbf{\#Ann. / sample} & \textbf{Train} & \textbf{Dev} & \textbf{Test} & \textbf{Total}\\ 
    \midrule
        Afrikaans (\textbf{\texttt{afr}}) &  Speeches  & 3 & 3 &  1,325 & 115 & 1,247 & 2,687 \\
        Algerian Arabic  (\textbf{\texttt{arq}}) &  Literature  & 10 & 4 to 9 & 1,686 & 182 & 1,674 & 3,542 \\
        Moroccan Arabic  (\textbf{\texttt{ary}})     & News, social media & 3 & 3 & 1,813 & 300 & 931 & 3,044 \\
        Chinese  (\textbf{\texttt{chn}})   &  Social media & 7 & 5 & 3,316 & 250 & 3,345 & 6,911 \\
        German  (\textbf{\texttt{deu}})     &   Social media& 10 & 7 & 3,700 & 294 & 3,690 & 7,684 \\
        English  (\textbf{\texttt{eng}}) &     Social media & 122 & 5 to 30 & 4,574 & 189 & 4,509 & 9,272 \\
        Latin American Spanish (\textbf{\texttt{esp}}) &     Social media & 12 & 5 & 2,835 & 256 & 2,340 & 5,431 \\
        Hausa  (\textbf{\texttt{hau}}) &   News, social media & 5 & 5 & 2,656 & 440 & 1,352 & 4,448 \\
        Hindi  (\textbf{\texttt{hin}}) &    Created & 5 & 4 to 5 & 2,841 & 108 & 1,070 & 4,019 \\
        Igbo  (\textbf{\texttt{ibo}}) &    News, social media & 3 & 3 & 2,988 & 497 & 1,502 & 4,987 \\
        Indonesian  (\textbf{\texttt{ind}}) &   Social media & 16 & 3 & -- & 247 & 1,409 & 1,656 \\
        Javanese  (\textbf{\texttt{jav}}) &    Social media & 13 & 3 & -- &  250 & 1,395 & 1,645 \\
        Kinyarwanda  (\textbf{\texttt{kin}}) &    News, social media & 3 & 3 & 5,350 & 426 & 1,298 & 4,299 \\
        Marathi  (\textbf{\texttt{mar}}) &      Created  & 4 & 4 &  2,590 & 108 & 1,103 & 3,864 \\
        Nigerian-Pidgin (\textbf{\texttt{pcm}}) &      News, social media & 3 & 3 & 1,553 & 888 & 2,691 & 8,929 \\
        Portuguese (Brazilian; \textbf{\texttt{ptbr}}) & Social media  & 5 & 5 & 2,318 & 228 & 2,580 & 5,398 \\
        Portuguese (Mozambican; \textbf{\texttt{ptmz}})  & News, social media & 3 & 3 & 2,995 & 258 & 780 & 2,591 \\
        Romanian (\textbf{\texttt{ron}})   & Social media & 8 & 3 to 8 & 1,352 & 123 & 1,893 & 4,536 \\
        Russian (\textbf{\texttt{rus}})   & Social media & 10 & 3 to 10 & 3,443 & 225 & 1,127 & 4,347 \\
        Sundanese (\textbf{\texttt{sun}})   & Social media & 16 & 3 & 1,495 & 292 & 1,351 & 2,995 \\
        Swahili (\textbf{\texttt{swa}})   & News, social media & 3 & 3 & 1,000 & 573 & 1,727 & 5,743 \\
        Swedish (\textbf{\texttt{swe}})   & Social media & 3 & 3 &  2,527 & 253 & 1,514 & 3,262 \\
        Tatar (\textbf{\texttt{tat}})   & Social media & 3 & 2 & 1,558 & 200 & 1,000 & 2,200 \\
        Ukrainian (\textbf{\texttt{ukr}})   & Social media & 106 & 5 & 3,133 & 255 & 2,278 & 5,060 \\
        Emakhuwa (\textbf{\texttt{vmw}})   & News, social media & 3 & 3 & 1,558 & 259 & 781 & 2,598 \\
        isiXhosa (\textbf{\texttt{xho}})   & News, social media & 3 & 3 & -- &  745 & 1,744 & 2,489 \\
        Yoruba (\textbf{\texttt{yor}})   & News & 3 & 3 & 3,133 & 520 & 1,572 & 5,225 \\
        isiZulu (\textbf{\texttt{zul}})   & News, social media & 3 & 3 & -- & 940 & 2,202 & 3,142 \\
    \bottomrule
    \end{tabular}
    }
    \caption{\textbf{Data sources, number of annotators, and data split sizes for the \datasetname~ datasets}, sorted alphabetically by language code. Datasets without training splits (–) were used exclusively for testing (see Section~\ref{sec:experiments}).}
    \label{tab:data_sources}
\end{table*}


As our \datasetname~ collection includes datasets in 28 different languages, curated and annotated by fluent speakers, we use several data sources, collection, and annotation strategies depending on 1)\ the availability of the textual data potentially rich in emotions and 2)\ access to annotators. We detail the choices made when selecting and balancing sources, annotating the instances, and controlling for data quality in the following section.

\subsection{Data Sources}
Selecting appropriate data can be challenging when resources are scarce. Therefore, we typically combine multiple sources, as shown in Table \ref{tab:data_sources}. Below, we outline the main textual domain inputs used to construct \datasetname.

\paragraph{Social media posts}
We use social media data collected from various platforms, including Reddit (e.g., \texttt{eng}, \texttt{deu}), YouTube (e.g., \texttt{esp}, \texttt{ind}, \texttt{jav}, \texttt{sun}), Twitter (e.g., \texttt{hau}, \texttt{ukr}), and Weibo (e.g., \texttt{chn}). For some languages, we re-annotate existing sentiment datasets for emotions (e.g., the sentiment analysis benchmark AfriSenti \cite{muhammad-etal-2023-afrisenti} for \texttt{ary}, \texttt{hau}, \texttt{kin}; the Twitter dataset by \citet{bobrovnyk2019automated} for \texttt{ukr}; the RED--v2 dataset \citep{ciobotaru-etal-2022-red} for \texttt{ron}).

\paragraph{Personal narratives, talks, and speeches}
Anonymised sentences from personal diary posts are ideal for extracting sentences where the speaker is centering their own emotions as opposed to the emotions of someone else. Hence, we use these in \texttt{eng}, \texttt{deu}, and \texttt{ptbr}, mainly from subreddits such as, e.g., IAmI.

Similarly, the \texttt{afr} dataset includes sentences from speeches and talks which constitute a good source for potentially emotive text.

\paragraph{Literary texts}
We manually translated the novel \textit{``La Grande Maison''} (The Big House) by the Algerian author Mohammed Dib \footnote{\url{https://en.wikipedia.org/wiki/La_Grande_Maison}} from French to Algerian Arabic and further post-processed the translation to generate sentences to be annotated by native speakers. Note that the translator is bilingual and a native Algerian Arabic speaker. 
Such a source is typically rich in emotions as it includes interactions between various characters. Moreover, Algerian Arabic is mainly spoken due to the Arabic diglossia, which makes this resource valuable since it highly differs from social media datasets in \texttt{arq}.

\paragraph{News data} 
Although we prefer emotionally rich social media data from different platforms, such data is not always available. Therefore, when data sources are limited, to collect a larger number of instances, we annotate news data and headlines in some African languages (e.g., \texttt{yor}, \texttt{hau}, and \texttt{vmw}).

\paragraph{Human-written and machine generated data}%
We create a dataset from scratch for Hindi (\texttt{hin}) and Marathi (\texttt{mar}). We ask annotators to generate emotive sentences on a given topic (e.g., family). In addition, we automatically translate a small section of the Hindi dataset to Marathi, and native speakers manually fix the translation errors. Finally, we augment both datasets with a few hundred quality-approved instances generated by ChatGPT.

\subsection{Pre-processing and Quality Control}
Prior to annotation, we preprocess the data by removing duplicates, invisible characters, garbled encoding, and incorrectly rendered emoticons. We anonymise all texts and exclude content with excessive expletives or dehumanising language.

\subsection{Annotating \datasetname}

As a text snippet can elicit multiple emotions simultaneously, we ask the annotators to select all the emotions that apply to a given text rather than choosing a single dominant emotion class.
The set of labels includes six categories of perceived emotions: \textit{anger, sadness, fear, disgust, joy, surprise}, and is considered \textit{neutral} if no emotion is picked. The annotators further rate the selected emotion(s) on a four-point intensity scale: 0 (no emotion), 1 (low intensity), 2 (moderate intensity level), and 3 (high intensity). We provide the definitions of the categories and annotation guide in \Cref{app:reliability}.

We use Amazon Mechanical Turk to annotate the English dataset, and Toloka\footnote{\url{https://toloka.ai}} to label the Russian, Ukrainian, and Tatar instances. However, as traditional crowdsourcing platforms do not have a large pool of annotators who speak various low-resource languages, we directly recruit fluent speakers to annotate the data and use the academic version of LabelStudio \cite{LabelStudio} and Potato \cite{pei2022potato} to set up our annotation platform.

\subsection{Annotators' Reliability}
While both inter-annotator agreement (IAA) and reliability scores evaluate annotation quality, they capture different aspects. IAA evaluates the extent to which annotators agree with one another, whereas reliability scores measure the consistency of aggregated labels across repeated annotation trials \citep{bws-naacl2016}. Consequently, reliability scores tend to increase with a larger number of annotations, while IAA scores do not depend on the number of annotations per instance.

We report the annotation reliability using Split-Half Class Match Percentage (SHCMP; \citealp{mohammad-2024-worrywords}). SHCMP extends the concept of Split-Half Reliability (SHR), traditionally applied to continuous scores \citep{bws-naacl2016}, to discrete categories, such as our emotion intensity labels.
SHCMP measures the extent to which $n$ bins (i.e., random subsets) classify items consistently. The dataset is randomly split into $n$ bins (corresponding to halves when $n=2$) $1,000$ times, and the proportion of instances receiving the same class label across bins is averaged to return the final SHCMP score. A higher SHCMP indicates greater reliability, meaning that repeated annotations would likely result in similar class labels. Additional details are provided in Appendix \ref{app:reliability}.
\Cref{fig:shcmp_heatmap} shows a heatmap of SHCMP scores for the \datasetname~ datasets. Overall, the SHCMP scores are high (greater than 60\% for $n=2$), indicating that our annotations are reliable.

\begin{figure}[t!]
    \centering
    \includegraphics[width=0.98\linewidth, trim={0.5cm 0 0 0}]{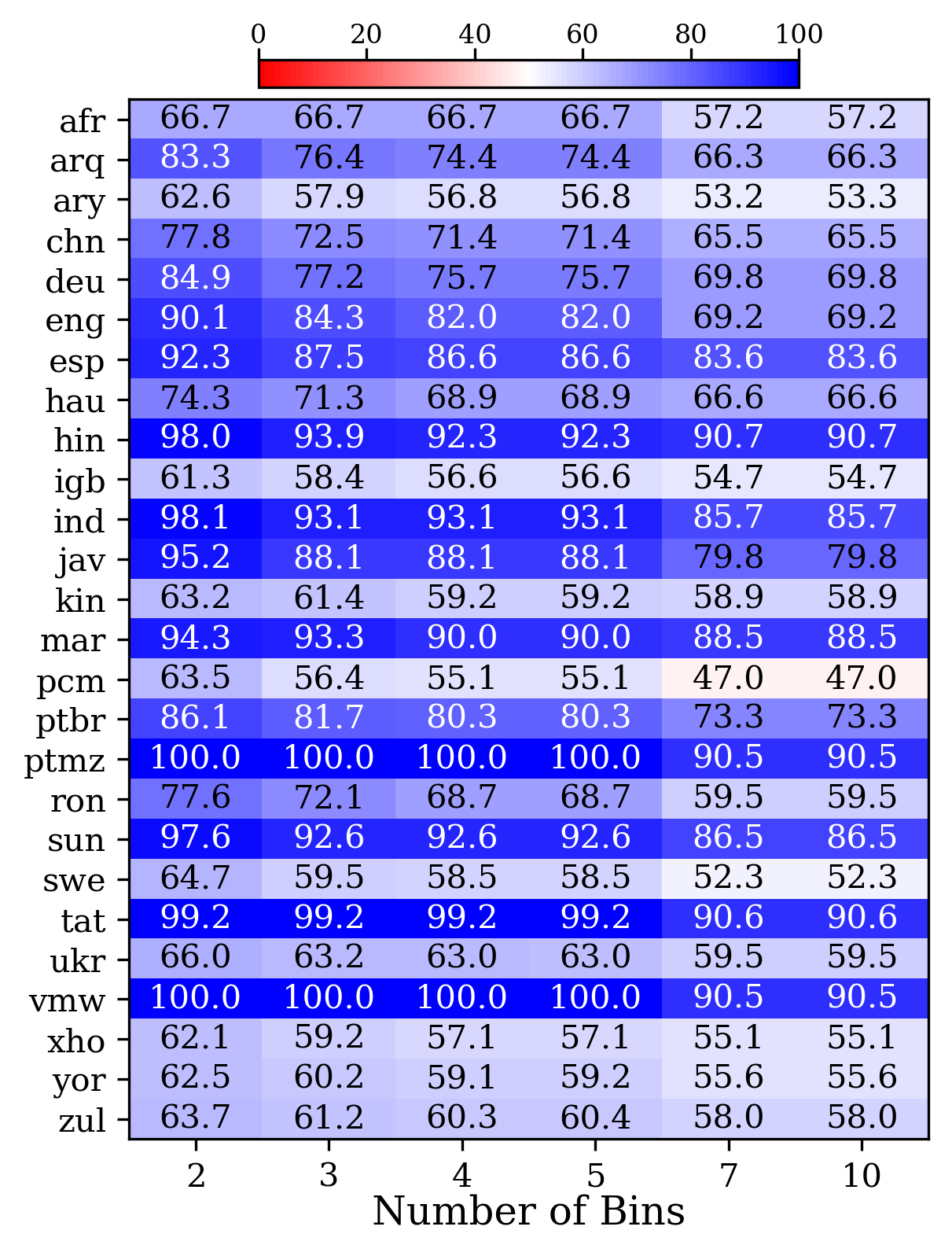}
    \caption{ \textbf{Split-Half Class Match Percentage, SHCMP (\%) values for the \datasetname~ datasets} across varying numbers of bins ($2$ to $10$). Higher values indicate better reliability scores. Note that \texttt{ptmz} and \texttt{vwm} have the same score as \texttt{vwm} instances were translated from \texttt{ptmz} and the translation was verified.}
    \label{fig:shcmp_heatmap}
\end{figure}

\subsection{Determining the Final Labels}

We expected a level of disagreement as emotions are complex, subtle, and perceived differently even from people within the same culture. In addition, text-based communication is limited as it lacks cues such as tone, relevant context, and information about the speaker. Our approach for aggregating the per-annotator emotion and intensity labels is detailed below. We also publicly share the individual (non-aggregated) annotations, recognising that annotator disagreement can provide useful signals in itself \citep{plank2022problem}. 

\paragraph{Aggregating the emotion labels}

The final emotion labels are determined based on the emotions and associated intensity values selected by the annotators. That is, the given emotion is considered present if: 
 \begin{enumerate}[noitemsep,nolistsep] 
    \item At least two annotators select a label with an intensity value of $1$, $2$, or $3$ (low, medium, or high, respectively).
    \item The average score exceeds a predefined threshold \(T\). We set \(T\) to \(0.5\).
\end{enumerate}

\paragraph{Aggregating the intensity labels} 

Once the labels for perceived emotions are assigned, we determine the final intensity score for each instance by averaging the selected intensity scores and rounding them up to the nearest integer. We assign intensity scores only for datasets in which the majority of instances are annotated by $\geq 5$ annotators, to ensure robustness.
Therefore, \datasetname~ includes emotion labels for 28 languages and intensity labels for 10 languages.

\subsection{Final Data Statistics}

\begin{figure*}[h]
    \centering
    \includegraphics[width=\linewidth]{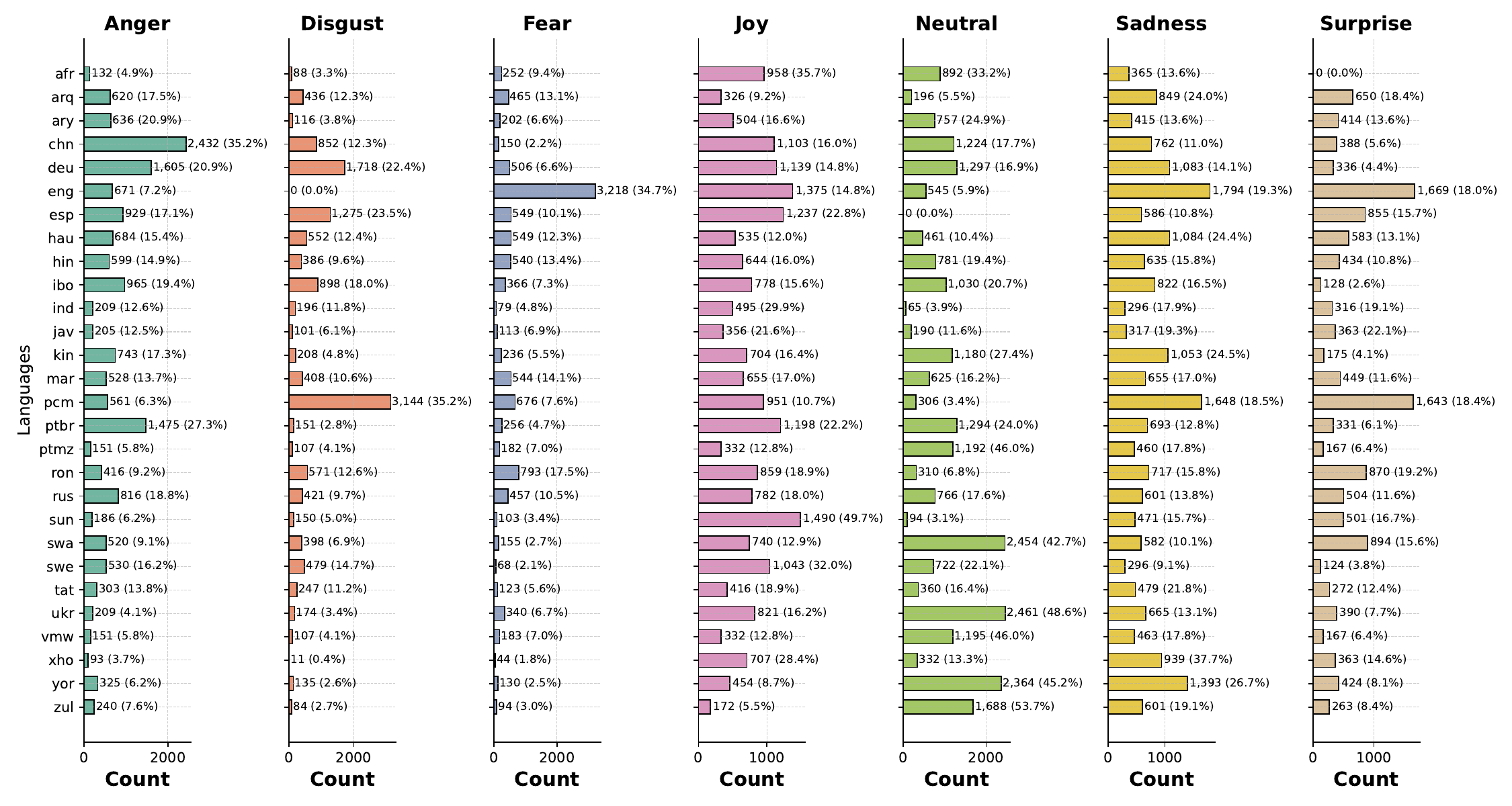}
    \caption{\textbf{Emotion label distribution across the \datasetname~datasets}. Each bar represents the number of labeled instances per emotion (i.e., anger, disgust, fear, joy, sadness, surprise, and neutral) and its percentage.}
    \label{fig:distr-label}
\end{figure*}

\Cref{fig:distr-label} shows the distribution of the annotated emotions in the \datasetname~datasets. The neutral class contains instances that do not belong to any of the six predefined categories (i.e., anger, disgust, fear, sadness, joy, and surprise). Although most languages include all six categories, the English dataset does not include disgust, and the Afrikaans one does not include surprise due to an insufficient class representation. Furthermore, class distributions show substantial variation as we chose various data sources as shown in Table \ref{tab:data_sources}.

\section{Experiments}\label{sec:experiments}

\subsection{Setup}

\begin{table*}[t]
\centering
\renewcommand{\arraystretch}{1.2}
\resizebox{\textwidth}{!}{%
\begin{tabular}{lccccc|ccccc}
\toprule
 & \multicolumn{5}{c|}{\textbf{Few-Shot Multi-Label Classification}} & \multicolumn{5}{c}{\textbf{Crosslingual Multi-Label Classification}} \\
\cmidrule(lr){2-6} \cmidrule(lr){7-11}
\textbf{Lang.} & \textbf{Qwen2.5-72B} & \textbf{Dolly-v2-12B} & \textbf{Llama-3.3-70B} & \textbf{Mixtral-8x7B} & \textbf{DeepSeek-R1-70B} & \textbf{LaBSE} & \textbf{RemBERT} & \textbf{XLM-R} & \textbf{mBERT} & \textbf{mDeBERTa} \\
\midrule
\texttt{afr} & 60.18 & 23.58 & \bestmono \textbf{61.28} & 53.69 & 43.66 & 35.12 & 35.04 & \bestcross \textbf{41.66} & 16.95 & 33.25 \\
\texttt{arq} & 37.78 & 38.59 & \bestmono \textbf{55.75} & 45.29 & 50.87 & \bestcross\textbf{35.93} & 33.78 & 35.87 & 31.38 & 35.92 \\
\texttt{ary} & \bestmono \textbf{52.76} & 24.27 & 44.96 & 35.07 & 47.21 & \bestcross \textbf{42.83} & 35.46 & 33.88 & 24.83 & 36.28 \\
\texttt{chn} & \bestmono \textbf{55.23} & 27.52 & 53.36 & 44.91 & 53.45 & 45.28& 24.56 & \bestcross \textbf{53.84} & 21.61 & 42.41 \\
\texttt{deu} & \bestmono \textbf{59.17} & 26.86 & 56.99 & 51.20 & 54.26 & 42.45 & 46.84 & \bestcross \textbf{47.26} & 28.60 & 42.61 \\
\texttt{eng} & 55.72 & 42.60 & \bestmono \textbf{65.58} & 58.12 & 56.99 & 36.71 & 37.54 & \bestcross \textbf{37.60} & 18.80 & 35.30 \\
\texttt{esp} & 72.33 & 36.41 & 61.27 & 65.72 & \bestmono \textbf{73.29} & 54.56 & \bestcross \textbf{57.37} & 44.52 & 30.09 & 37.09 \\
\texttt{hau} & 43.79 & 29.43 & 50.91 & 40.40 & \bestmono \textbf{51.91} & \bestcross \textbf{38.46} & 31.98 & 16.69 & 15.59 & 32.80 \\
\texttt{hin} & \bestmono \textbf{79.73} & 27.59 & 60.59 & 62.19 & 76.91 & 69.78 & 13.75 & \bestcross\textbf{69.96} & 36.94 & 57.74 \\
\texttt{ibo} & \bestmono \textbf{37.40} & 24.31 & 33.18 & 31.90 & 32.85 & \bestcross\textbf{18.13} & 7.49 & 10.42 & 9.94 & 9.52 \\
\texttt{ind} & \bestmono \textbf{57.29} & 36.61 & 39.20 & 54.37 & 49.51 & \bestcross\textbf{47.50} & 37.64 & 25.39 & 26.87 & 35.68 \\
\texttt{jav} & \bestmono \textbf{50.47} & 36.18 & 41.88 & 48.37 & 43.05 & 46.24 & \bestcross\textbf{46.38} & 20.39 & 26.16 & 35.34 \\
\texttt{kin} & 31.96 & 19.73 & \bestmono \textbf{34.36} & 26.35 & 32.52 & \bestcross\textbf{30.35} & 18.38 & 13.12 & 20.90 & 17.30 \\
\texttt{mar} & 74.58 & 25.69 & 67.40 & 50.36 & \bestmono\textbf{76.68} & 74.65 & \bestcross\textbf{77.24} & 76.21 & 42.32 & 54.05 \\
\texttt{pcm} & 38.66 & 34.41 & \bestmono\textbf{48.67} & 45.61 & 45.00 & \bestcross\textbf{33.29} & 1.01 & 21.08 & 22.55 & 25.39 \\
\texttt{ptbr} & \bestmono \textbf{51.60} & 25.90 & 45.03 & 41.64 & 51.49 & 41.51 & 41.84 & \bestcross\textbf{43.09} & 23.86 & 34.42 \\
\texttt{ptmz} & \bestmono \textbf{40.44} & 16.70 & 34.06 & 36.52 & 39.58 & \bestcross\textbf{31.44} & 29.67 & 7.30 & 13.54 & 24.46 \\
\texttt{ron} & 68.18 & 43.58 & \bestmono \textbf{71.28} & 68.51 & 65.02 & 69.79 & \bestcross\textbf{76.23} & 65.21 & 61.50 & 60.60 \\
\texttt{rus} & 73.08 & 29.72 & 62.61 & 61.72 & \bestmono\textbf{76.97} & 61.32 & \bestcross\textbf{70.43} & 21.14 & 37.15 & 29.70 \\
\texttt{sun} & 42.67 & 32.20 & \bestmono \textbf{46.33} & 42.10 & 44.61 & \bestcross\textbf{34.79} & 19.43 & 25.92 & 25.29 & 27.31 \\
\texttt{swa} & 27.36 & 17.63 & 29.47 & 26.51 & \bestmono\textbf{33.27} & 21.66 & \bestcross\textbf{18.99}& 16.94 & 18.61 & 14.94 \\
\texttt{swe} & 48.89 & 21.79 & \bestmono\textbf{50.26} & 48.61 & 44.60 & 44.24 & \bestcross\textbf{51.18} & 10.08 & 28.86 & 43.28 \\
\texttt{tat} & 51.58 & 25.12 & 49.84 & 39.44 & \bestmono\textbf{53.86} & \bestcross\textbf{60.66} & 44.54 & 39.58 & 35.81 & 47.72 \\
\texttt{ukr} & \bestmono\textbf{54.76} & 17.16 & 42.34 & 40.15 & 51.19 & 44.37 & \bestcross\textbf{49.56} & 34.06 & 25.69 & 35.12 \\
\texttt{vmw} & \bestmono\textbf{20.41} & 16.03 & 18.96 & 19.00 & 19.09 & 9.65 & 5.22 & \bestcross\textbf{12.66} & 12.11 & 11.74 \\
\texttt{xho} & 29.56 & 24.12 & 30.79 & 22.92 & 29.08 & 31.39 & 12.73 & 11.48 & 17.08 & 22.86 \\
\texttt{yor} & 24.99 & 16.00 & 23.70 & 19.67 & \bestmono\textbf{27.44} & \bestcross\textbf{11.64} & 5.33 & 6.64 & 9.62 & 10.03 \\
\texttt{zul} & 22.03 & 14.72 & 21.48 & 20.38 & 20.38 & 18.16 & 15.26 & 10.92 & 13.04 & 13.87 \\
\midrule
\textbf{AVG} & \bestmono\textbf{49.71} & 26.88 & 47.12 & 43.56 & 49.21 & \bestcross \textbf{40.50} & 33.63 & 30.61 & 24.16 & 32.38 \\
\bottomrule
\end{tabular}
}
\caption{\textbf{Average F1-Macro for multi-label emotion classification}. In the few-shot setting, we predict the emotion class on test set in 28 languages. In the crosslingual setting, we train on all languages within a language family except the target language, and evaluate on the test set of the target language. The best performance scores in few-shot and crosslingual settings are highlighted in \textcolor{blue}{blue} and \textcolor{orange}{orange}, respectively.}
\label{tab:mono_cross_lingual_result}
\end{table*}
We report the data split sizes in \Cref{tab:data_sources}. The test sets are relatively large, ranging from approximately $1,000$ to nearly $3,000$ instances. Datasets without training data are excluded from training and are used solely for testing in cross-lingual settings.

For our baseline experiments, we evaluate multi-label emotion classification and emotion intensity prediction using both Multilingual Language Models (MLMs) and Large Language Models (LLMs).

\paragraph{Multi-label emotion classification in few-shot settings} \label{para:few-shot}
We report emotion classification performance using five LLMs--\texttt{Qwen2.5-72B}~\citep{yang2024qwen2}, \texttt{Dolly-v2-12B}~\citep{DatabricksBlog2023DollyV2}, \texttt{LlaMA-3.3-70B}~\citep{touvron2023llama}, \texttt{Mixtral-8x7B}~\citep{jiang2024mixtral}, and \texttt{DeepSeek-R1-70B}~\citep{deepseekai2025deepseekr1incentivizingreasoningcapability}. We prompt the LLMs using Chain-of-Thought (CoT) reasoning to predict the presence of each emotion from the predefined set. We set the number of few-shot examples to 8 and consider only the first generated answer (i.e., top-1). We report macro F1 scores across $28$ languages. In \Cref{app:multilingual}, we also provide monolingual classification results for the $24$ languages with training data (see \Cref{tab:monolingual_result_mlm}).

\paragraph{Multi-label emotion classification in crosslingual settings}
We report the macro F-score results for systems trained without using any data in the $28$ target languages when testing on each. Hence, we train MLMs on all languages in one family (see \Cref{fig:lgge_fams}) except for one held-out target language, which we test on and report the results for each test set. For families with only one language (i.e., Sino-Tibetan, Creole, and Turkic), we train on Slavic languages (\texttt{rus} and \texttt{ukr}) and test on \texttt{tat}; two Niger-Congo languages (\texttt{swa} and \texttt{yor}) and test on \texttt{pcm}; and on \texttt{rus} and test on \texttt{chn}.

\paragraph{Emotion intensity prediction}
 We report Pearson correlation scores for systems trained on the intensity-labeled training sets in $10$ languages.



\begin{table*}[t]
\centering
\renewcommand{\arraystretch}{1.2}
\resizebox{\textwidth}{!}{%
\begin{tabular}{lccccc|ccccc}
\toprule
 & \multicolumn{5}{c|}{\textbf{Multilingual Language Models (MLMs)}} & \multicolumn{5}{c}{\textbf{Large Language Models (LLMs)}} \\
\cmidrule(lr){2-6} \cmidrule(lr){7-11}
\textbf{Lang.} & \textbf{LaBSE} & \textbf{RemBERT} & \textbf{XLM-R} & \textbf{mBERT} & \textbf{mDeBERTa} & \textbf{Qwen2.5-72B} & \textbf{Dolly-v2-12B} & \textbf{Llama-3.3-70B} & \textbf{Mixtral-8x7B} & \textbf{DeepSeek-R1-70B} \\
\midrule
\texttt{arq} & 1.42 & \bestmlm\textbf{1.64} & 0.89 & 1.10 & 0.47 & 29.54 & 3.80 & 36.29 & 31.05 & \bestllm\textbf{36.37} \\
\texttt{chn} & 23.37 & \bestmlm\textbf{40.53} & 36.92 & 21.96 & 23.25 & 46.17 & 8.11 & \bestllm\textbf{51.86} & 46.52 & 48.57 \\
deu & 28.93 & \bestmlm\textbf{56.21} & 38.30 & 17.35 & 18.14 & 43.30 & 7.43 & 53.46 & 47.60 & \bestllm\textbf{54.78} \\
\texttt{eng} & 35.34 & \bestmlm\textbf{64.15} & 37.36 & 25.74 & 8.85 & \bestllm\textbf{55.99} & 13.35 & 44.14 & 55.26 & 48.08 \\
\texttt{esp} & 56.89 & \bestmlm\textbf{72.59} & 55.72 & 27.94 & 29.18 & 51.11 & 10.49 & 51.64 & 55.54 & \bestllm\textbf{60.74} \\
\texttt{hau} & 26.13 &\bestmlm\textbf{27.03} & 24.68 & 2.79 & 0.00 & 27.00 & 6.43 & \bestllm\textbf{39.16} & 25.84 & 38.85 \\
\texttt{ptbr} & 20.62 & \bestmlm\textbf{29.74} & 18.24 & 8.36 & 1.32 & 38.20 & 9.02 & 40.90 & 39.17 & \bestllm\textbf{46.72} \\
\texttt{ron} & 35.57 & \bestmlm\textbf{55.66} & 37.77 & 21.99 & 4.63 & 55.48 & 12.62 & 45.87 & 57.07 & \bestllm\textbf{57.69} \\
\texttt{rus} & 68.43 & \bestmlm\textbf{87.66}& 68.96 & 37.63 & 5.03 & 58.25 & 13.96 & 57.56 & 56.01 & \bestllm\textbf{62.28} \\
\texttt{ukr} & 13.75 & \bestmlm\textbf{39.94} & 36.16 & 4.32 & 3.51 & 37.74 & 6.04 & 36.99 & 38.74 & \bestllm\textbf{43.54} \\
\midrule
\textbf{AVG} & 30.54 & \bestmlm\textbf{46.61} & 35.25 & 16.35 & 9.97 & 43.03 & 8.74 & 45.78& 43.97 & \bestllm\textbf{48.88} \\
\bottomrule
\end{tabular}
}
\caption{\textbf{Pearson correlation scores for intensity classification} using MLMs and LLMs. The best performance scores are highlighted in \textcolor{blue}{blue} and \textcolor{orange}{orange}, respectively.
}
\label{tab:merged_results_track_b}
\end{table*}






\subsection{Experimental Results}


\Cref{tab:mono_cross_lingual_result} reports the results of few-shot and crosslingual experiments for multi-label emotion classification and 
 \Cref{tab:merged_results_track_b} reports those for emotion intensity classification. Our results corroborate how challenging emotion classification is for LLMs, even for high-resource languages such as \texttt{eng} and \texttt{deu}. The performance is worse for low-resource languages, for which \texttt{Dolly-v2-12B} performs the worst, and \texttt{Qwen2.5-72B} performs the best on average.
 
We observe the largest performance for \texttt{yor} with a maximum of $27.44$. \texttt{hin}, \texttt{mar}, and \texttt{tat} have the best performance among all languages, which is unsurprising since the \texttt{tat} dataset is single-labeled, and close to 70\% and 80\% of the test data for \texttt{mar} and \texttt{hin} respectively are single-labeled. 

\paragraph{Multi-label emotion recognition results} 
The crosslingual experiments demonstrate that model performance depends on both the languages used for transfer learning and those included in the pretraining of the models. For instance, in some cases, training on languages from the same family improves performance and even surpasses few-shot settings, e.g., \texttt{swe} benefits when RemBERT is fine-tuned on other Germanic languages. However, all Niger-Congo languages, particularly \texttt{vmw}, benefit the least from crosslingual transfer across all models, with RemBERT performing the worst. This is largely due to the severe under-resourcedness of these languages, even when data is combined.
Notably, XLM-R performs exceptionally well on languages such as \texttt{deu}, \texttt{chn}, \texttt{hin}, and \texttt{ptbr}, but struggles significantly with others (e.g., \texttt{swe}, \texttt{ptmz}). In contrast, mDeBERTa yields the most consistent results across most languages, even though it shows low performance on \texttt{ibo}, \texttt{vmw}, and \texttt{yor}, which are not part of the CC-100 corpus \cite{conneau-etal-2020-unsupervised} used in its training. While mDeBERTa was also not trained on \texttt{arq}, the inclusion of Modern Standard Arabic (MSA) in its pretraining data might have positively influenced its performance.

Overall, our results indicate that multilingual models transfer more effectively to languages seen during pretraining, while often producing random or unreliable outputs for languages absent from their training data.

\paragraph{Emotion intensity prediction} 
For intensity detection, a more challenging task, \texttt{Dolly-v2-12B} performs the worst, whereas \texttt{DeepSeek-R1-70B} shows promising results, outperforming other models in most languages. \texttt{Llama-3.3-70B} and \texttt{Qwen2.5-72B} achieve the highest scores in English.
Interestingly, MLMs tend to perform better on high-resource languages--RemBERT, in particular, achieves strong results for \texttt{deu}, \texttt{eng}, \texttt{esp}, and \texttt{rus}, with \texttt{chn} being the only exception. In contrast, for primarily spoken, low-resource vernaculars (e.g., \texttt{arq}), LLMs demonstrate striking improvements --\texttt{DeepSeek-R1-70B}, for instance, achieves improvements exceeding $36$ points.


\section{Analysis}
\begin{figure*}[ht]
    \centering
    \begin{subfigure}[t]{0.3\textwidth}
         \centering
         \includegraphics[width=\linewidth]{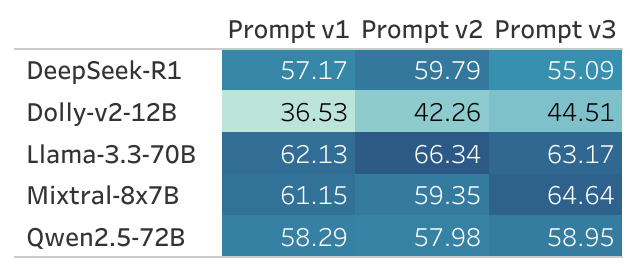}
         \caption{\textbf{Performance of different LLMs} across three prompt paraphrases on the English test set. Different prompts impact model performance.
         }
         \label{fig:prompt_variant}
     \end{subfigure}
      \hfill
        \begin{subfigure}[t]{0.32\textwidth}
       \centering
         \includegraphics[width=\linewidth]{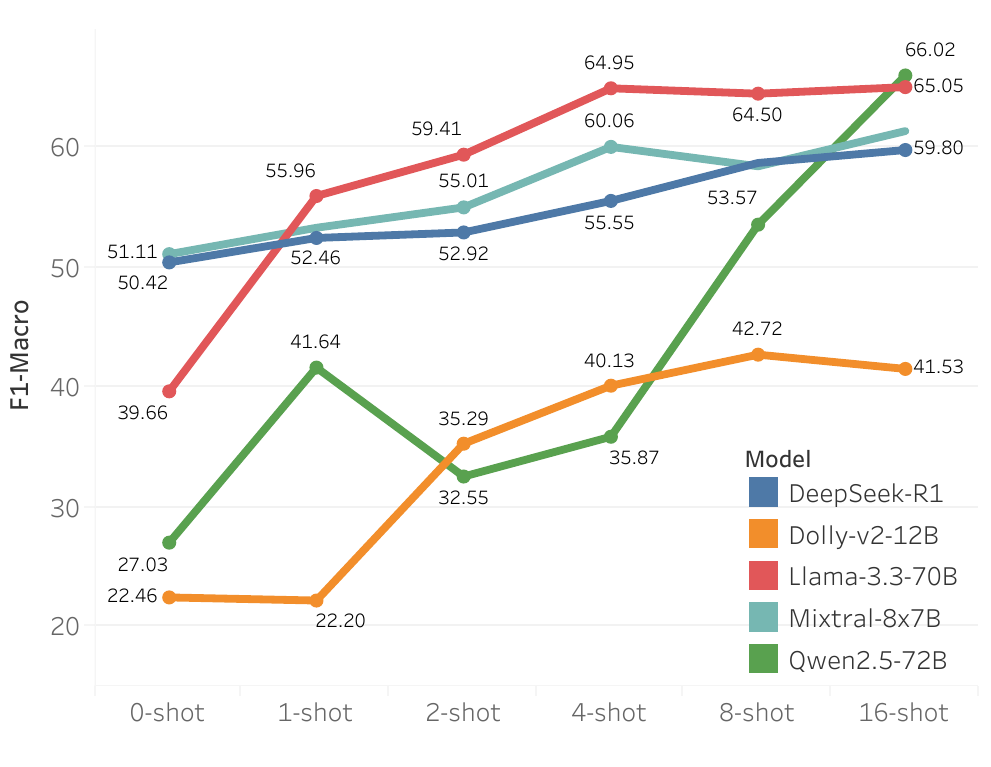}
         \caption{\textbf{Few-shot performance of LLMs} on the English test set.
         Performance improves with more shots.
         }
         \label{fig:few_shot_performance}
       \end{subfigure}
        \hfill
      \begin{subfigure}[t]{0.32\textwidth}
       \centering
           \includegraphics[width=\linewidth]{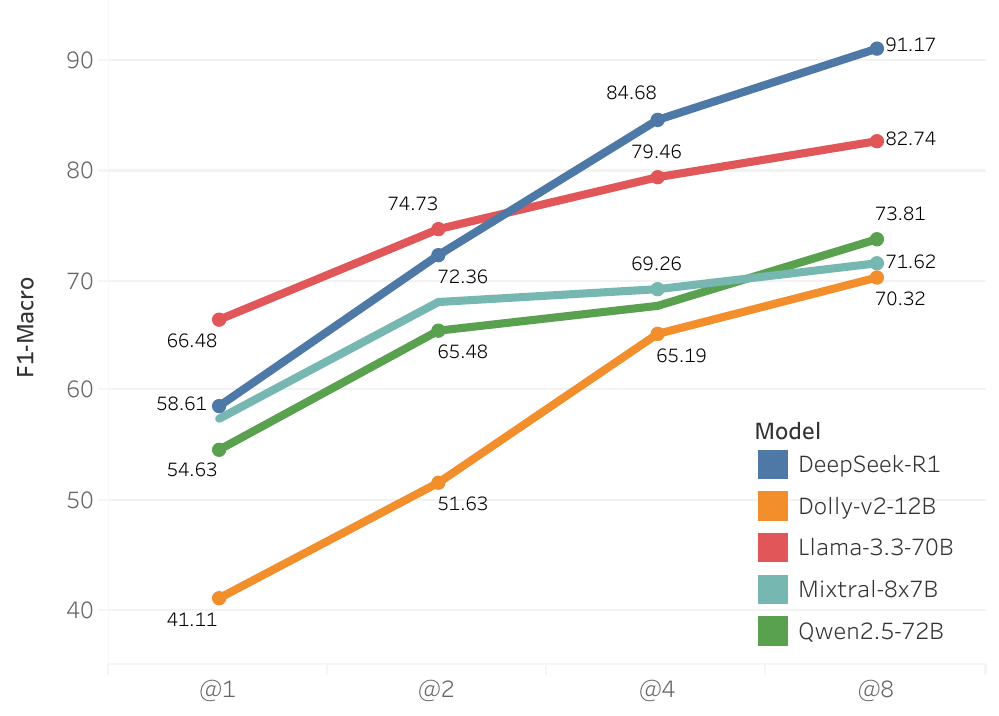}
         \caption{\textbf{Pass@k performance of different LLMs} on the English test set. Higher $k$ values increase the likelihood of retrieving the correct answer.
         }
         \label{fig:topk_performance}
       \end{subfigure}
     \caption{Ablation studies on the effect of prompt wording variation, few-shot examples, and pass@k predictions conducted on the English test set.}
\end{figure*}

The results in \Cref{fig:prompt_variant} suggest that LLM performance is highly dependent on the prompt wording when asking for the presence of emotion on the English test set using different paraphrases of the same text.
Further, Figure \ref{fig:few_shot_performance} shows that, when testing the effect of n-shot settings on the English test set, we observe a significant improvement in performance with more shots, with \texttt{Mixtral-8x7B} and \texttt{Llama-3.3-70B} outperforming other models. However, the scores tend to reach a plateau at $4$ shots for all LLMs except for \texttt{Qwen2.5-72B}, which suggests that $4$ to $8$ shots may be sufficient to obtain stable results.
 In addition, when testing how likely we can get the correct answer when prompting LLMs to generate tokens based on a selection of $k$ generations, the results shown in Figure \ref{fig:topk_performance} suggest that increasing the value of $k$ results consistently in better performance, particularly when using \texttt{DeepSeekR1-70B}, which achieves an F-score $>90$ when $k=8$ whereas \texttt{Mixtral-8x7B} shows a smaller change in performance followed by \texttt{Llama-3.3-70B} and \texttt{Qwen2.5-72B}. The ranking of the models for $k=8$ remains consistent with the one achieved for $k=1$.



\begin{figure}
    \centering
    \includegraphics[width=1\linewidth]{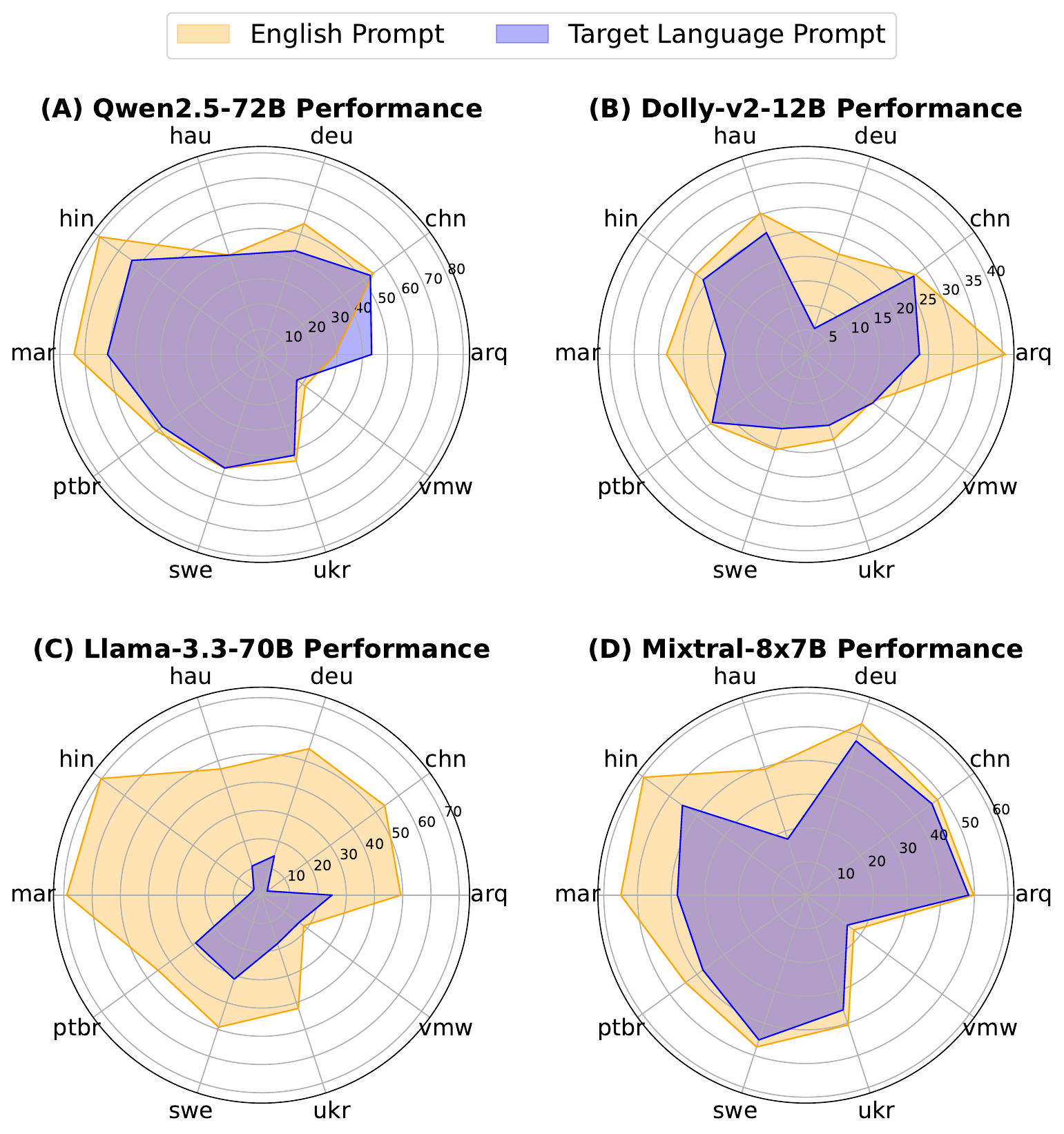}
    \caption{\textbf{Comparing models' performance across languages} when prompted in English (\textcolor{orange}{orange}) vs. when prompted in the target language (\textcolor{blue}{blue}). LLMs perform better when prompted in English.}
    
    \label{fig:radar_chart}
\end{figure}
When comparing the performance of models prompted in English versus the target language, \Cref{fig:radar_chart} shows that LLMs generally perform better with English prompts except for \texttt{arq}, where \texttt{Qwen2.5-72B} achieves better results when prompted in Modern Standard Arabic (MSA). The improvement from using English prompts is particularly evident in low-resource languages (e.g., \texttt{hau}, \texttt{mar}, \texttt{vmw}), where models like \texttt{Dolly-v2-12B} and \texttt{Llama-3.3-70B} perform poorly with prompts in the target language.

\section{Related Work}

Appraisal theories of emotion propose that emotions arise from our evaluation of events based on personal experiences, leading to different emotional responses among individuals \citep{arnold1960emotion, frijda1986emotions, lazarus1991emotion, scherer2009dynamic, ellsworth2013appraisal, moors2013appraisal, roseman2013appraisal, ortony2022cognitive}. The theory of constructed emotion claims that emotions are not hard-wired or universal, but rather conceptual constructs formed by the brain \citep{barrett2016, barrett2017emotions}.

Early work in NLP primarily focused on sentiment analysis--identifying whether a text conveys positive, negative, or neutral valence \citep{Mohammad2016, muhammad-etal-2023-semeval}. More recent research has shifted toward a broader goal: detecting specific emotions in text, such as anger, fear, joy, and sadness. This shift aligns with discrete models of emotion, including Paul Ekman’s six basic emotions \citep{ekman1992there} and Plutchik’s Wheel of Emotions \citep{Plutchik1980}, which includes anger, disgust, fear, happiness, sadness, surprise, anticipation, and trust.

Several initiatives have created emotion classification datasets for languages other than English such as Italian \citep{bianchi-etal-2021-feel}, Romanian \citep{ciobotaru-etal-2022-red}, Indonesian \citep{saputri2018emotion}, and Bengali \citep{iqbal2022bemoc}. However, the field remains predominantly Western-centric. Although multilingual datasets such as XED \citep{ohman2020xed} and XLM-EMO \citep{bianchi2022xlm} exist, the latter’s reliance on translated data for over ten languages may not adequately reflect cultural nuances in emotional expression. Emotions are culture-sensitive and highly contextual, shaped by different norms and values \citep{hershcovich-etal-2022-challenges, havaldar-etal-2023-multilingual, mohamed-etal-2024-culture, plaza-del-arco-etal-2024-emotion}.

Furthermore, although emotions can co-occur \citep{vishnubhotla-etal-2024-emotion-granularity}, most existing datasets assume a single-label classification framework. While GoEmotions \citep{demszky-etal-2020-goemotions} addresses multi-label emotion classification, to our knowledge, no multilingual resources capture simultaneous emotions and intensity across languages.
This work aims to advance the field by introducing emotion-labeled data for 28 languages. Given the lack of consensus around what constitutes a low-resource language, approximately 15 to 17 among these could reasonably be considered as such.

\section{Conclusion}
We presented \datasetname, a collection of emotion recognition datasets in 28 languages spoken across various continents. The instances in \datasetname~are multi-labeled, collected, and annotated by fluent speakers, with 10 datasets annotated for emotion intensity. When testing LLMs on our dataset collection, the results show that they still struggle with predicting perceived emotions and their intensity levels, especially for under-resourced languages. Further, our results show that LLM performance is highly dependent on the wording of the prompt, its language, and the number of shots in few-shot settings.
We publicly release \datasetname, our annotation guidelines, and individual labels to the research community.

\section*{Limitations} \label{sec:limitations}
Emotions are subjective, subtle, expressed, and perceived differently. We do not claim that \datasetname~covers the true emotions of the speakers, is fully representative of the language use of the 28 languages, or covers all possible emotions. We discuss this extensively in the Ethics Section. 

We are aware of the limited data sources in some low-resource languages. Therefore, our datasets cannot be used for tasks that require a large amount of data from a given language. However, they remain a good starting point for research in the area.

\section*{Ethical Considerations} \label{sec:ethics}
Emotion perception and expression are inherently subjective and nuanced, as they are closely tied to a myriad of factors (e.g., cultural background, social group, personal experiences, and social context).
As such, it is impossible to determine with absolute certainty how someone is feeling based solely on short text snippets. Therefore, we explicitly state that our datasets focus on \textit{perceived} emotions--that is, the emotions most people believe the speaker may have felt.
Accordingly, we do not claim to annotate the \textit{true} emotion of the speaker, as this cannot be definitively inferred from short texts alone.
We recognise the importance of this distinction, as perceived emotions may differ from actual emotions.

We also acknowledge potential biases in our data. Text-based communication inherently carries biases, and our data sources may reflect such tendencies. Similarly, annotators may come with their own subtle, internalised biases. Moreover, although many of our datasets focus on low-resource languages, we do not claim they fully capture the usage of these languages. While we took care to exclude inappropriate content, some instances may have been inadvertently overlooked.

We strongly encourage careful ethical reflection before using our datasets. Use of the data for commercial purposes or by state actors in high-risk applications is strictly prohibited unless explicitly approved by the dataset creators. Systems developed using our datasets may not be reliable at the individual instance level and are sensitive to domain shifts. They should not be used to make critical decisions about individuals, such as in health-related applications, without appropriate expert oversight. See \citet{mohammad2022ethics, mohammad-2023-best} for a comprehensive discussion on these issues.

Finally, all annotators involved in the study were compensated at rates exceeding the local minimum wage.

\section*{Acknowledgments}
Shamsuddeen Muhammad acknowledges the support of Google DeepMind and Lacuna Fund, an initiative co-founded by The Rockefeller Foundation, Google.org, and Canada’s International Development Research Centre. The views expressed herein do not necessarily represent those of Lacuna Fund, its Steering Committee, its funders, or Meridian Institute.
\newline
Nedjma Ousidhoum would like to thank Abderrahmane Samir Lazouni, Lyes Taher Khalfi, Manel Amarouche, Narimane Zahra Boumezrag, Noufel Bouslama, Abderraouf Ousidhoum, Sarah Arab, Wassil Adel Merzouk, Yanis Rabai, and another annotator who would like to stay anonymous for their work and insightful comments. 
\newline
Idris Abdulmumin gratefully acknowledges the ABSA UP Chair of Data Science for funding his post-doctoral research and providing compute resources.
\newline
Jan Philip Wahle, and Terry Ruas, Bela Gipp, Florian Wunderlich were partially supported by the Lower Saxony Ministry of Science and Culture and the
VW Foundation.
\newline
Meriem Beloucif acknowledges Nationella Språkbanken (the Swedish National Language Bank) and Swe-CLARIN – jointly funded by the Swedish Research Council (2018–2024; dnr 2017-00626) and its 10 partner institutions for funding the Swedish annotations.
\newline
Rahmad Mahendra acknowledges the funding by Hibah Riset Internal Faculty of Computer Science, Universitas Indonesia under contract number: NKB-13/UN2.F11.D/HKP.05.00/2024 for supporting the annotation of the Indonesian, Javanese, and Sundanese datasets.
\newline
Alexander Panchenko would like to thank Nikolay Ivanov, Artem Vazhentsev, Mikhail Salnikov, Maria Marina, Vitaliy Protasov, Sergey Pletenev, Daniil Moskovskiy, Vasiliy Konovalov, Elisey Rykov, and Dmitry Iarosh for their help with the annotation for Russian. Preparation of Tatar data was funded by AIRI and completed by Dina Abdullina, Marat Shaehov, and Ilseyar Alimova.
\newline
Yi Zhou would like to thank Gaifan Zhang, Bing Xiao, and Rui Qin for their help with the annotations and for providing feedback.
\newline
Daryna Dementieva and Nikolay Babakov would like to acknowledge Toloka.ai for the research annotation grant. Daryna Dementieva would like as well to acknowledge Alexander Fraser, the head of Data Statistics and Analytics Group at Technical University of Munich, for the support.





\bibliography{anthology,custom}

\appendix

\clearpage

\section{PLMs and LLMs Used}
\subsection{PLMs}
 \begin{enumerate}
    \item \url{https://huggingface.co/google-bert/bert-base-multilingual-cased}
    \item \url{https://huggingface.co/FacebookAI/xlm-roberta-large}
    \item \url{https://huggingface.co/microsoft/mdeberta-v3-base}
    \item \url{https://huggingface.co/sentence-transformers/LaBSE}
    \item \url{https://huggingface.co/microsoft/infoxlm-large}
    \item \url{https://huggingface.co/google/rembert}
\end{enumerate}

\subsection{LLMs}
\begin{enumerate}
    \item \url{https://huggingface.co/databricks/dolly-v2-12b}
    \item \url{https://huggingface.co/meta-llama/Meta-Llama-3.3-70B}
    \item \url{https://huggingface.co/Qwen/Qwen2.5-72B-Instruct}
    \item \url{https://huggingface.co/mistralai/Mixtral-8x7B-Instruct-v0.1}
    \item \url{https://huggingface.co/deepseek-ai/DeepSeek-R1-Distill-Llama-70B}
\end{enumerate}



\section{Data sources}
\begin{itemize}
    \item \textbf{\texttt{afr}}: Speeches from \citet{barnard2014nchlt}.
    \item \textbf{\texttt{arq}}: Manually translated novel (\textit{La Grande Maison} by the Algerian author Mohammed Dib).
    \item \textbf{\texttt{ary, hau, ibo, kin, pcm, swa, xho, zul}}: \textbf{Afrisenti} \citet{muhammad-etal-2023-afrisenti} and BBC news headlines.
    \item \textbf{\texttt{chn}}: Weibo dataset \url{https://github.com/aoguai/WeiboHotListDataSet?tab=readme-ov-file}.
    \item \textbf{\texttt{deu}}: Anonymised Reddit data from nine German-language subreddits: \textit{de, einfach\_posten, FragReddit, beziehungen, schwanger, de\_IAmA, germany, depression\_de, Lagerfeuer}.
    \item \textbf{\texttt{eng}}: Personal narratives from the AskReddit subreddit collected by \citet{ouyang2015modeling} and instances from \citet{zhuang2024my}.
    \item \textbf{\texttt{esp}}: YouTube comments from Latin American (i.e., Ecuadorian, Colombian, and Mexican) channels across three genres: \textit{News/Politics, Entertainment, Education}.
    \item \textbf{\texttt{hin, mar}}: Newly created emotion dataset. Most instances were manually drafted, while some were generated using ChatGPT.
    \item \textbf{\texttt{ind, jav, sun}}: YouTube comments from Indonesian videos.
    \item \textbf{\texttt{ron}}: Data from the subreddit \textit{r/Romania}, YouTube, and tweets from \citet{ciobotaru-etal-2022-red}.
    \item \textbf{\texttt{rus}}: Russian Twitter corpus \url{https://study.mokoron.com}.
    \item \textbf{\texttt{swe}}: Sentiment dataset from the Swedish data bank \cite{SvenskABSAbank}.
    \item \textbf{\texttt{tat}}: Instances from \citet{krylova2016languages}.
    \item \textbf{\texttt{vmw, ptmz}}: News headlines from \citet{ali2024building}.
    \item \textbf{\texttt{yor}}: News data from BBC \yoruba and Alaroye. \url{https://alaroye.org/}.
\end{itemize}

\section{Annotation}
\subsection{Annotation Guidelines and Definitions}\label{app:guide}

This is a guide for annotating text for emotion classification. The purpose of this study is to analyze the emotions expressed in a text. It is important to note that emotions can often be inferred even if they are not explicitly stated. 

\paragraph{Task}
The task involves classifying text into predefined emotion categories. The annotated dataset will be used for training emotion classification models and studying how emotions are conveyed through language.
\paragraph{Emotion Categories}
We categorize emotions into the following seven classes:

\noindent \textbf{Joy}
\begin{itemize}
    \item Definition: Expressions of happiness, pleasure, or contentment.
    \item  Example: \textit{"I just passed my exams!"}
\end{itemize}

\noindent \textbf{Sadness}
\begin{itemize}
    \item Definition: Expressions of unhappiness, sorrow, or disappointment.
    \item Example: \textit{"I miss my family so much. It's been a tough year."}
\end{itemize}

\noindent \textbf{Anger}
\begin{itemize}
    \item Definition: Expressions of frustration, irritation, or rage.
    \item Example:\textit{"Why is the internet so slow today?!"}
\end{itemize}

\noindent \textbf{Fear}
\begin{itemize}
    \item Definition: Expressions of anxiety, apprehension, or dread.
    \item Example: \textit{"There's a huge storm coming our way. I hope everyone stays safe."}
\end{itemize}

\noindent \textbf{Surprise}
\begin{itemize}
    \item Definition: Expressions of astonishment or unexpected events.
    \item Example: \textit{"I can't believe he just proposed to me!"}
\end{itemize}

\noindent \textbf{Disgust}
\begin{itemize}
    \item Definition: A reaction to something offensive or unpleasant.
    \item Examples: \textit{"That video was sickening to watch."}
\end{itemize}

\noindent \textbf{Neutral}
\begin{itemize}
    \item Definition: Texts that do not express any of the above emotions.
    \item Example: \textit{"The weather today is sunny with a chance of rain."}
\end{itemize}

\vspace{5pt}
\textbf{Note:} Factual statements can indicate an emotional state without explicitly stating it. For example:
\begin{itemize}
    \item \textit{"An earthquake today killed hundreds of people in my home town."}
\end{itemize}

\noindent Surprise differs from joy in that it represents an unexpected event, which may or may not be associated with happiness.

\vspace{10pt}

\paragraph{Emotion Description Categories}
The following list provides a broader categorization of emotions by including synonyms and related emotional states.

\vspace{5pt}

\noindent \textbf{Anger}
\begin{itemize}
    \item Includes: \textit{irritated, annoyed, aggravated, indignant, resentful, offended, exasperated, livid, irate}, etc.
\end{itemize}

\noindent \textbf{Sadness}
\begin{itemize}
    \item Includes: \textit{melancholic, despondent, gloomy, heartbroken, longing, mourning, dejected, downcast, disheartened, dismayed}, etc.
\end{itemize}

\noindent \textbf{Fear}
\begin{itemize}
    \item Includes: \textit{frightened, alarmed, apprehensive, intimidated, panicky, wary, dreadful, shaken}, etc.
\end{itemize}

\noindent \textbf{Happiness}
\begin{itemize}
    \item Includes: \textit{joyful, elated, content, cheerful, blissful, delighted, gleeful, satisfied, ecstatic, upbeat, pleased}, etc.
\end{itemize}

\noindent \textbf{Surprise}
\begin{itemize}
    \item Includes: \textit{taken aback, bewildered, astonished, amazed, startled, stunned, shocked, dumbstruck, confounded, stupefied}, etc.
\end{itemize}

\noindent \textbf{Joy}
\begin{itemize}
    \item Includes: \textit{happiness, delight, elation, pleasure, excitement, cheerfulness, bliss, euphoria, contentment, jubilation}.
\end{itemize}

\subsubsection{Emotion Intensity} \label{sec:examples}

After selecting the emotion category, annotators were further asked to select the intensity label, which could be: 0: No Emotion, 1 - Slight Emotion,  2: Moderate Emotion and 3: High Emotion. The following examples illustrate different levels of emotion intensity.

\vspace{10pt}

\noindent \textbf{Anger}
\begin{itemize}
    \item No Anger: \textit{"I walked through the empty streets, the quiet hum of the city like a distant whisper."}
    \item Slight Anger: \textit{"The buzz of voices around me blended into a monotonous drone, failing to distract from the pang of annoyance at the delay."}
    \item High Anger: \textit{"When his friend's brother knocked on the door, he was greeted with a shotgun blast through the door, which left him dead at the doorstep."}
\end{itemize}

\vspace{5pt}

\subsection{Pilot Annotation}
We run a pilot annotation on different languages to further refine our guidelines. This has mainly led to clarifications related to the labeling process. For instance, the annotators were reminded that they should select all the labels that apply for a given text snippet, and that one label can encompass more than one specific emotion (e.g., in \texttt{arq}, we explained that a complex perceived emotion such as bitterness or jealousy might involve both anger and sadness).

\subsection{ Formula for Determining Final Labels}

\paragraph{Aggregating emotion labels}
Aggregating emotion labels can be formally expressed as:

\tcbset{
  colback=gray!6,
  colframe=black,
  sharp corners,
  width=\linewidth,
  boxrule=0.2mm
}
\begin{tcolorbox}
\resizebox{\linewidth}{!}{$
L_{\text{final}} =
\begin{cases} 
1, & \text{if } \text{Count}(1, 2, 3) \geq 2 \text{ and } \text{AvgScore} > T, \\
0, & \text{otherwise}.
\end{cases}
$}

\vspace{1mm}

\resizebox{\linewidth}{!}{$
\text{Count}(1, 2, 3) = \sum_{i=1}^N \mathbb{1}(A_i \in \{1, 2, 3\}), \quad
\text{AvgScore} = \frac{1}{N} \sum_{i=1}^N A_i
$}
\end{tcolorbox}
Where:
\begin{itemize}
    \item \(A_i\) is the rating provided by annotator \(i\).
    \item \(N\) is the total number of annotators.
    \item \(\mathbb{1}(A_i \in \{1, 2, 3\})\) Membership function that returns 1 if \(A_i \in \{1, 2, 3\}\), and 0 otherwise.
    \item \(T\) is the threshold for the average score, which we set as \(T = 0.5\)
\end{itemize}

\paragraph{Aggregating intensity}

Aggregating intensity can be formally expressed as:

 \tcbset{colback=gray!6, colframe=black, sharp corners, width=\linewidth, boxrule=0.2mm}

\begin{tcolorbox}
\[
\text{AvgScore} = \frac{\sum_{i=1}^N A_i}{N},
\]

\[
L_{\text{final}}=
\begin{cases} 
0, & \text{if } 0 \leq \text{AvgScore} < 1, \\
1, & \text{if } 1 \leq \text{AvgScore} < 2, \\
2, & \text{if } 2 \leq \text{AvgScore} < 3, \\
3, & \text{if } \text{AvgScore} = 3.
\end{cases}
\]

\end{tcolorbox}

Where:
\begin{itemize}
    \item \(A_i\) is the intensity score provided by annotator \(i\), where \(A_i \in \{0, 1, 2, 3\}\).
    \item \(N\) is the total number of annotators.
\end{itemize}

\section{SCHMP Calculation}\label{app:reliability}

The computation of SHCMP involves the following steps:

\paragraph{1. Random Splitting with Tie-Breaking} 
The dataset of $N$ annotated items is randomly divided into two equal subsets, $A_1$ and $A_2$. For datasets with an odd number of annotations, probabilistic tie-breaking is applied to ensure balanced splits.

\paragraph{2. Class Assignment} 
For each item $x_i \, (i = 1, 2, \ldots, N)$:
\begin{itemize}
    \item Assign $x_i$ a score based on its annotations in $A_1$ and $A_2$.
    \item Let $C_1(x_i)$ and $C_2(x_i)$ denote the class of $x_i$ derived from $A_1$ and $A_2$, respectively.
\end{itemize}

\paragraph{3. Class Binning} 
To manage continuous scores, divide the range of possible scores $[-3, 3]$ into equal-sized bins, where the bin size $b$ is determined as:
\[
b = \frac{6}{\#\text{Bins}}.
\]
Scores from $A_1$ and $A_2$ are then assigned to their respective bins, denoted as $c_1$ and $c_2$.

\paragraph{4. Match Calculation} 
Define a match indicator $M(x_i)$ to evaluate consistency for each item:
\[
M(x_i) = 
\begin{cases} 
1, & \text{if } |c_1 - c_2| < 1, \\
0, & \text{otherwise.}
\end{cases}
\]
This ensures that items are considered consistent if their scores fall into the same bin or adjacent bins.

\paragraph{5. Proportion of Matches} 
Compute the total number of matches, $N_{\text{match}}$, across all items:
\[
N_{\text{match}} = \sum_{i=1}^{N} M(x_i).
\]

\paragraph{6. SHCMP Computation} 
The SHCMP score is calculated as the proportion of matches, expressed as a percentage:
\[
\text{SHCMP (\%)} = \frac{N_{\text{match}}}{N} \times 100.
\]

\paragraph{7. Averaging} 
We repeat the process $k$ times with different random splits and compute the average SHCMP score:
\[
\text{SHCMP}_{\text{final}} = \frac{1}{k} \sum_{j=1}^{k} \text{SHCMP}_j,
\]
where $\text{SHCMP}_j$ is the SHCMP score from the $j$-th split.

\begin{table*}[htbp]
\centering
\begin{tabular}{lccc|ccc|ccc}
    \hline
    \multirow{2}{*}{Language} & \multicolumn{3}{c|}{Train Set (\%)} & \multicolumn{3}{c|}{Development Set (\%)} & \multicolumn{3}{c}{Test Set (\%)} \\
    & Single & Multi & Neutral & Single & Multi & Neutral & Single & Multi & Neutral \\
    \hline
    \texttt{chn} & 54.00 & 23.74 & 22.26 & 53.60 & 23.58 & 22.82 & 53.90 & 24.30 & 21.80 \\
    \texttt{sun} & 58.94 & 36.18 & 4.88 & 59.09 & 36.26 & 4.65 & 59.40 & 36.07 & 4.54 \\
    \texttt{afr} & 47.79 & 6.69 & 45.52 & 56.14 & 7.86 & 36.01 & 37.39 & 10.35 & 52.26 \\
    \texttt{swe} & 43.16 & 16.60 & 40.24 & 46.30 & 20.37 & 33.33 & 42.76 & 18.81 & 38.43 \\
    \texttt{swa} & 41.67 & 3.33 & 55.00 & 45.78 & 3.56 & 50.66 & 46.26 & 3.81 & 49.93 \\
    \texttt{esp} & 61.02 & 38.98 & 0.00 & 65.22 & 34.78 & 0.00 & 65.14 & 34.86 & 0.00 \\
    \texttt{arq} & 28.53 & 50.05 & 9.42 & 28.57 & 50.00 & 10.71 & 27.95 & 44.76 & 8.35 \\
    \texttt{ptbr} & 52.11 & 13.80 & 34.09 & 61.06 & 11.82 & 27.12 & 52.68 & 13.59 & 33.73 \\
    \texttt{ptmz} & 52.00 & 0.44 & 47.56 & 50.92 & 0.37 & 48.71 & 53.03 & 0.51 & 46.45 \\
    \texttt{ukr} & 44.77 & 2.24 & 52.99 & 47.24 & 2.36 & 50.39 & 45.23 & 1.79 & 52.98 \\
    \texttt{mar} & 67.69 & 8.56 & 23.75 & 68.57 & 7.62 & 23.81 & 68.94 & 9.33 & 21.73 \\
    \texttt{rus} & 64.63 & 11.08 & 24.29 & 66.35 & 12.23 & 21.42 & 66.91 & 12.89 & 20.20 \\
    \texttt{ibo} & 72.44 & 3.63 & 23.93 & 61.12 & 10.91 & 27.97 & 73.61 & 3.97 & 22.42 \\
    \texttt{amh} & 50.82 & 27.68 & 21.50 & 56.13 & 30.31 & 16.56 & 48.50 & 24.67 & 26.83 \\
    \texttt{deu} & 41.78 & 34.05 & 24.17 & 41.84 & 35.19 & 22.97 & 41.23 & 32.10 & 26.66 \\
    \texttt{vmw} & 52.80 & 0.45 & 46.75 & 53.49 & 0.39 & 46.12 & 53.46 & 0.52 & 46.32 \\
    \texttt{pcm} & 55.00 & 40.46 & 4.54 & 50.00 & 36.63 & 4.37 & 51.57 & 38.08 & 4.35 \\
    \texttt{eng} & 38.64 & 47.02 & 14.34 & 34.07 & 42.22 & 9.70 & 38.58 & 48.76 & 10.34 \\
    \texttt{hin} & 66.35 & 10.80 & 22.85 & 60.40 & 7.92 & 31.68 & 77.31 & 5.66 & 13.92 \\
    \texttt{tat} & 81.48 & 0.00 & 18.52 & 84.00 & 0.00 & 16.00 & 85.71 & 0.00 & 14.29 \\
    \hline
\end{tabular}
\caption{Percentage distribution of \textit{SingleLabel}, \textit{MultiLabel}, and \textit{NeutralLabel} for the Train, Development, and Test Sets.}
\label{tab:all_sets_label_percentage}
\end{table*}

\begin{table*}[t]
\centering
\renewcommand{\arraystretch}{1.1}
\setlength{\tabcolsep}{4pt} 
\resizebox{0.5\linewidth}{!}{%
\begin{tabular}{lccccc}
\toprule
 & \multicolumn{5}{c}{\textbf{Monolingual Multi-Label Classification}} \\
\cmidrule(lr){2-6}
\textbf{Lang.} & LaBSE & RemBERT & XLM-R & mBERT & mDeBERTa \\
\midrule
\texttt{afr} &  30.76 &  \bestmono \textbf{37.14} & 10.82 & 25.87 & 16.66 \\
\texttt{arq} & \bestmono \textbf{45.46} & 41.41 & 31.98 & 41.75 & 29.68 \\
\texttt{ary} & 45.81 & \bestmono \textbf{47.16} & 40.66 & 36.87 & 38.00 \\
\texttt{chn} & 53.47 & 53.08 & \bestmono  \textbf{58.48} & 49.61 & 44.47 \\
\texttt{deu} & 55.02 & \bestmono  \textbf{64.23} & 55.37 & 46.78 & 44.09 \\
\texttt{eng} & 64.24 & \bestmono \textbf{70.83} & 67.30 & 58.26 & 58.94 \\
\texttt{esp} & 72.88 & \bestmono  \textbf{77.44} & 29.85 & 54.41 & 60.17 \\
\texttt{hau} & 58.49 & \bestmono \textbf{59.55} & 36.95 & 47.33 & 48.59 \\
\texttt{hin} & 75.25 & \bestmono  \textbf{85.51} & 33.71 & 54.11 & 54.34 \\
\texttt{ibo} & 45.90 & \bestmono \textbf{47.90} & 18.36 & 37.23 & 31.92 \\
\texttt{ind} & -- & -- & -- & -- & -- \\
\texttt{jav} & -- & -- & -- & -- & -- \\
\texttt{kin} & \bestmono \textbf{50.64} & 46.29 & 32.93 & 35.61 & 38.00 \\
\texttt{mar} & 80.76 & \bestmono \textbf{82.20} & 78.95 & 60.01 & 66.01 \\
\texttt{pcm} & 51.30 & \bestmono \textbf{55.50} & 52.03 & 48.42 & 46.21 \\
\texttt{ptbr} & \bestmono \textbf{42.60} & 42.57 & 15.40 & 32.05 & 24.08 \\
\texttt{ptmz} & 36.95 & \bestmono \textbf{45.91} & 30.72 & 14.81 & 21.89 \\
\texttt{ron} & 69.79 & \bestmono  \textbf{76.23} & 65.21 & 61.50 & 60.60 \\
\texttt{rus} & 75.62 & \bestmono \textbf{83.77} & 78.76 & 61.81 & 54.79 \\
\texttt{sun} & 36.93 & \bestmono  \textbf{37.31} & 19.66 & 27.88 & 21.65 \\
\texttt{swa} & \bestmono \textbf{27.53} & 22.65 & 22.71 & 22.99 & 22.84 \\
\texttt{swe} & 49.23 & \bestmono \textbf{51.98} & 34.63 & 44.24 & 40.90 \\
\texttt{tat} & \bestmono \textbf{57.71} & 53.94 & 26.48 & 43.49 & 35.02 \\
\texttt{ukr} & 50.07 & \bestmono \textbf{53.45} & 17.77 & 31.74 & 28.55 \\
\texttt{vmw} & \bestmono \textbf{21.13} & 12.14 & 9.92 & 10.28 & 11.13 \\
\texttt{xho} & -- & -- & -- & -- & -- \\
\texttt{yor} & \bestmono \textbf{32.55} & 9.22 & 11.94 & 21.03 & 17.88 \\
\texttt{zul} & -- & -- & -- & -- & -- \\
\bottomrule
\end{tabular}
}
\caption{\textbf{Average F1-Macro for monolingual multi-label emotion classification.} Each model is trained and evaluated within the same language. The best results are highlighted in \textcolor{blue}{blue}.}
\label{tab:monolingual_result_mlm}
\end{table*}
\label{app:multilingual}

\section{Experimental Settings} \label{sec:appendix}

For LLMs, we used the default parameters from HuggingFace except for temperature which we set to 0 for deterministic output and top-k is set to 1. Only for the top-k ablations in which top-k > 1 in \Cref{fig:topk_performance}, we set temperature to 0.7. We ask all LLMs to perform CoT. We trained on the train set for 2 epochs with a learning rate of 1e-5 and and evaluated on test set. For MLMs experiments, we trained on the training set for 2 epochs with a learning rate of 1e-5 and evaluated on the test set.

\begin{table*}
\centering
\begin{tabular}{p{3cm} p{10.5cm}}
\toprule
\textbf{Prompt Version} & \textbf{Prompt Text} \\
\midrule
Prompt v1 & 
\texttt{Evaluate whether the following text conveys the emotion of \{\{EMOTION\}\}.\newline
Think step by step before you answer.\newline
Finish your response with 'Therefore, my answer is ' followed by 'yes' or 'no':\newline\newline\{\{INPUT\}\}} \\\\
Prompt v2 &
\texttt{Analyze the text below for the presence of \{\{EMOTION\}\}.\newline
Explain your reasoning briefly and conclude with 'Answer:' followed by either 'yes' or 'no'.\newline\newline\{\{INPUT\}\}} \\\\
Prompt v3 &
\texttt{Examine the following text to determine whether \{\{EMOTION\}\} is present.\newline
Provide a concise explanation for your assessment and end with 'Answer:' followed by either 'yes' or 'no'.\newline\newline\{\{INPUT\}\}} \\
\bottomrule
\end{tabular}
\caption{The prompt variants used in the monolingual emotion recognition ablation study.}
\label{tab:prompt-variants}
\end{table*}

\begin{figure*}[t]
    \begin{AIbox}{Track A: Example Few-Shot Prompt}
    \parbox[t]{\textwidth}{
        \textbf{\#\#\# Task: \#\#\#} \\
        Analyze the text below for the presence of anger. \\
        Explain your reasoning briefly and conclude with 'Answer:' followed by either 'yes' or 'no'. \\

        \textbf{\#\#\# Examples: \#\#\#} \\
        Example 1:\\
        Input: '''When I answered the phone, my heart beat extremely fast... I was very nervous!'''\\
        Answer: no \\

        Example 2:\\
        Input: '''I'll never forget how businesslike and calm the Israeli guy was.'''\\
        Answer: no \\

        Example 3:\\
        Input: '''I wake up, my eyes fluttering open to a shield of darkness.'''\\
        Answer: no \\

        Example 4:\\
        Input: '''I lay in a large bed, the sheets and quilt pulled up to my chin, and the curtains were drawn to keep out the light.'''\\
        Answer: no \\

        Example 5:\\
        Input: '''Either way that idiot is gone.'''\\
        Answer: yes \\

        Example 6:\\
        Input: '''Seriously... did I really just shut my finger in the car door.'''\\
        Answer: yes \\

        Example 7:\\
        Input: '''I was really uncomfortable because I was sitting behind my dad and there isn't enough room for my legs.'''\\
        Answer: yes \\

        Example 8:\\
        Input: '''He damn disturb plz, cover my head with a shirt that a customer which have body odour just tried on!!'''\\
        Answer: yes \\

        \par\noindent\rule{\textwidth}{0.4pt}

        \textbf{\#\#\# Your Turn: \#\#\#} \\
        Input: '''/ o \ So today I went in for a new exam with Dr. Polvi today, I had to file new paperwork for the automobile accident case which is being done differently than the scoliosis stuff. So he comes in and starts talking about insurance stuff and how this looks bad since I was getting treatment on my neck and stuff already blah blah.'''
    }
    \end{AIbox}
    \caption{Example of the few-shot prompt template for assessing anger in Track A.}
    \label{fig:prompt_example_tracka}
\end{figure*}

\begin{figure*}[t]
    \begin{AIbox}{Track B: Example Few-Shot Prompt}
    \parbox[t]{\textwidth}{
        \textbf{\#\#\# Task: \#\#\#} \\
        In this task, you will assess the level of anger in a given text (0 = none, 1 = low, 2 = medium, 3 = high). \\
        Summarize your reasoning and conclude with 'Answer:' followed by the correct number. \\

        \textbf{\#\#\# Examples: \#\#\#} \\
        Example 1:\\
        Input: '''I try extremely hard to keep my details hidden. It was nice to know that what I had given people to know was pleasant, but I couldn't deny the knot that was still in my stomach.'''\\
        Answer: 0 \\

        Example 2:\\
        Input: '''I knew we were almost there when my midwife's voice got more excited and Joey leaned in real close and said into my ear, `` Don't stop pushing! '' '''\\
        Answer: 0 \\

        Example 3:\\
        Input: '''One ended up going to prison.'''\\
        Answer: 1 \\

        Example 4:\\
        Input: '''Not to mention noisy.'''\\
        Answer: 1 \\

        Example 5:\\
        Input: '''" but Urban Dictionary confirmed Spook is indeed a racial slur.'''\\
        Answer: 2 \\

        Example 6:\\
        Input: '''And..at his funeral, they fired him!'''\\
        Answer: 2 \\

        Example 7:\\
        Input: '''I ended up metaphorically throwing my hands in the air in disgust and just cancelling my account altogether.'''\\
        Answer: 3 \\

        Example 8:\\
        Input: '''He would manipulate me into it and I was extremely upset.'''\\
        Answer: 3 \\

        \par\noindent\rule{\textwidth}{0.4pt}

        \textbf{\#\#\# Your Turn: \#\#\#} \\
        Input: '''So today I went in for a new exam with Dr. Polvi today, I had to file new paperwork for the automobile accident case which is being done differently than the scoliosis stuff. So he comes in and starts talking about insurance stuff and how this looks bad since I was getting treatment on my neck and stuff already blah blah.'''
    }
    \end{AIbox}
    \caption{Example of the few-shot prompt template for assessing anger in Track B.}
    \label{fig:prompt_example_trackb}
\end{figure*}

\end{document}